\documentclass{article}
\usepackage{caption} 
\usepackage{subcaption}
\usepackage{color}
\usepackage{hyperref}
\usepackage{mathtools}
\usepackage{amsmath}
\usepackage{amsfonts}
\usepackage{spconf,amsmath,graphicx}

\definecolor{green}{RGB}{0,128,0}
\definecolor{blue}{RGB}{0,0,255}
\definecolor{red}{RGB}{220,20,60}
\definecolor{orange}{RGB}{255,140,0}

\newlength{\bibitemsep}
\newlength{\bibparskip}
\let\oldthebibliography\thebibliography
\renewcommand\thebibliography[1]{%
  \oldthebibliography{#1}%
  \setlength{\parskip}{\bibitemsep}%
  \setlength{\itemsep}{\bibparskip}%
}

\title{Modelling Paralinguistic Properties in Conversational Speech to Detect Bipolar Disorder and Borderline Personality Disorder}
\name{Bo Wang$^1$$^,$$^3$, Yue Wu$^2$$^,$$^3$, Nemanja Vaci$^4$, Maria Liakata$^3$$^,$$^5$, Terry Lyons$^2$$^,$$^3$, Kate E A Saunders$^1$}
\address{
  $^1$Department of Psychiatry and $^2$Mathematical Institute, University of Oxford, UK\\
  $^3$The Alan Turing Institute, London, UK \\
  $^4$Department of Psychology, University of Sheffield, UK\\
  $^5$School of Electronic Engineering and Computer Science, Queen Mary University of London, UK\\
  \texttt{bo.wang@psych.ox.ac.uk}
  }
\begin{document}
%\ninept
%
\maketitle
\begin{abstract}
Bipolar disorder (BD) and borderline personality disorder (BPD) are two chronic mental health conditions that clinicians find challenging to distinguish based on clinical interviews, due to their overlapping symptoms. In this work, we investigate the automatic detection of these two conditions by modelling both verbal and non-verbal cues in a set of interviews. We propose a new approach of modelling short-term features with visibility-signature transform, and compare it with widely used high-level statistical functions. We demonstrate the superior performance of our proposed signature-based model. Furthermore, we show the role of different sets of features in characterising BD and BPD.
% we show different sets of features characterising BD and BPD respectively. 
\end{abstract}
\begin{keywords}
% mood disorder, spoken dialogue, computational paralinguistics, speech analysis, rough path signature
bipolar disorder, borderline personality disorder, speech analysis, paralinguistic modelling, path signature
\end{keywords}
\section{Introduction}
\label{sec:intro}
Bipolar disorder (BD) is a mood disorder characterised by extreme mood swings between manic highs and depressive lows that can last from days to weeks each. Borderline personality disorder (BPD) is a type of personality disorder marked by a long-term pattern of varying moods, self-image and behaviour. They both can seriously affect the patients' ability to function in work and social activities \cite{coryell1993enduring,goldberg1995course,zimmerman2015psychosocial}. They also can co-occur in 10\%-20\% of the cases and since the symptomatology of these two disorders is very similar, differentiating between them poses a diagnostic challenge for the clinicians \cite{zimmerman2013relationship,zimmerman2013distinguishing}. However, accurate diagnosis is crucial, as the most effective treatment for BD being pharmacological while for BPD psychotherapeutic. Standard diagnostic assessment for BD and BPD rely primarily on clinical interviews where the patients have to describe their accounts of symptoms. As a result it can be potentially influenced by patients' retrospective recall biases and cognitive limitations \cite{low2020automated}.

In recent years a number of machine learning based studies have investigated the use of acoustic features from speech for automated assessment across several psychiatric disorders showing diagnostic potentials \cite{low2020automated,cummins2015review,parola2020voice}. Existing work on bipolar disorder showed mood episodes affecting patients' speech and in turn acoustic and dialogue features extracted from speech
% both acoustic features \cite{faurholt2016voice,gideon2016mood,guidi2017features} and non-verbal features such as duration and number of pauses \cite{maxhuni2016classification,aldeneh2019identifying} 
can be used in detecting mood states \cite{faurholt2016voice,gideon2016mood,aldeneh2019identifying}. In this work, we aim to model both verbal and non-verbal cues in non-clinical interviews for the distinction between BD and BPD, which remains understudied.
% develop a method that automatically extracts both the verbal and non-verbal cues from non-clinical interviews for the distinction between BD and BPD, which remains understudied.
% However, different audio recording environments can affect the predictive capabilities of a system relying on acoustic features. \cite{gideon2016mood} examined the acoustic variations caused by recording with two different types of phones and the preprocessing needed for mood detection. While much of the existing work have focused on recognising mood episodes of BD patients either using clinical assessment calls \cite{gideon2016mood,aldeneh2019identifying} or personal phone calls \cite{faurholt2016voice}, the distinction between BD and BPD remains understudied. In this work, we aim to develop a method that automatically extracts both the verbal and non-verbal cues from non-clinical interviews for detecting BD and BPD. We also analyse the effect of the recording environment, by comparing models learnt from phone interviews to the ones conducted in the meeting room, as we hypothesise the dynamics of face-to-face conversation are significantly different to phone calls.

One key step of dialogue-based automated assessment system is modelling a series of conversational utterances and aggregate features from each segment to a fixed-size representation for down-stream classification or regression. Traditionally, a very popular approach has been to apply a set of high-level statistical functions  (e.g. mean, max, variance, linear regression coefficients, etc.) for feature aggregation \cite{faurholt2016voice,gideon2016mood,aldeneh2019identifying,matton2019into}, where the role of these aggregation functions is to describe the global characteristics of given spoken conversation. However, the conversational dynamics is not effectively modelled during this process and important sequential information may be ignored as a result. Recurrent neural networks (RNN) are designed to explicitly model sequential data,  
% But RNNs are not suitable for small clinical datasets as they generally require large amount of labelled training samples to avoid overfitting.
however, they are prone to overfit on small datasets. Signature transform, initially introduced in stochastic analysis, is a non-parametric approach of encoding sequential data and capturing the order information in the data. It has been proven effective in a range of machine learning tasks involving sequential modelling \cite{arribas2017signature,wang2019path,kidger2019deep}. A recent study \cite{wang2020learning} proposed summarising linguistic and turn-taking behaviour features from recorded interviews using signature. However, minimal description of its working was provided, and no comparative evaluation was given.
% In this paper, we perform comparative evaluation for high-level statistical functions (HSF) and signature transform (SIG) in feature aggregation to detect BD and BPD.

The contributions of this work are as follows: (1) We study three types of acoustic features for the classification of BD and BPD patients; 
% (2), We investigate the effect of acoustic properties of the interviews and the impact of recording environments; 
(2), We propose using visibility transform to enhance the signature representation of conversational speech; (3), We evaluate and compare between high-level statistical functions (HSF) and signature transform (SIG) based models, and show the superiority of SIG.

\section{AMoSS-I Dataset}
\label{sec:data}
% The Automated Monitoring of Symptoms Severity (AMoSS) study was a longitudinal study in which a smartphone app and a range of wearables were used for the daily self-report of mood \cite{tsanas2016daily}. During the study, 50 participants were interviewed and transcribed to gather qualitative feedback, resulting the AMoSS-Interview (AMoSS-I) dataset \cite{wang2020learning}. 
The Automated Monitoring of Symptoms Severity Interview (AMoSS-I) dataset \cite{wang2020learning} contains 50 participants, who were interviewed and transcribed to gather qualitative feedback of the original AMoSS study. Among these participants, 21 had a BD diagnosis, 17 had been diagnosed with BPD and 12 were healthy controls.
% All the diagnoses had been confirmed before the study. While the exclusion criteria for BD and BPD were comorbidity of each diagnosis, the exclusion criteria for the control group included history of neurological/psychiatric disorder and having a first degree relative with a history of BD or BPD \cite{mcgowan2019circadian}. 
Each participant was interviewed only once.

Among the 50 one-on-one qualitative interviews in AMoSS-I, 32 were recorded in the meeting room while the remaining 18 were phone interviews. 
% in which only the interviewer's end of the phone was recorded. The duration of the interviews are ranging from 7.7 to 47.9 minutes. 
As summarised in Table~\ref{tab:data2}, the phone interviews are more likely to be shorter than the ones conducted in the meeting room. We also take the difference of the peak and trough values for Root-mean-square (RMS) level measured over a 10 ms window, showing the difference of loudness within each audio recording. Comparing to the meeting room interviews, on average the phone interviews are shown to have significant larger difference of loudness within each one, mainly due to the way of recording. Additionally, we find the noise level in the room interviews to be higher as seen in their much lower signal-to-noise ratio of 21.16 dB, computed by Waveform Amplitude Distribution Analysis (WADA-SNR) \cite{kim2008robust}. 

\begin{table}[htb]
\begin{center}
\caption{Differences in sample size and acoustics between the room and phone interviews. $N$ is the number of the interviews. Audio length in minutes summarised in the form of the median +/- the interquartile range. The percentages of clipped samples are averaged over each set of interviews. RMS diff is the difference between the peak and trough values for RMS level measured over a 10 ms window, in dBFS. Signal-to-noise ratio (SNR) measures the amount of non-speech in a speech signal in decibels (dB).}
\label{tab:data2}
\begin{tabular}{c|c|c|c|c|c}
\hline \bf Env & $\bf{N}$ & \bf Length & \bf \%Clipped & \bf RMS diff & \bf SNR\\ \hline  %& \bf RMS
%  &  & \\
% \hline
Room & 32 & $23\pm{12}$ & 10.28\% & 64.91 & 21.16\\ % & 0.038
Phone & 18 & $19\pm{10}$ & 3.48\% & 83.10 & 95.30\\ % & 0.044
Both & 50 & $22\pm{11}$ & 7.83\% & 71.46 & 47.85\\ % & 0.040
%\%Clipped(median+IQR):$10.15\pm{2.30}$;$3.35\pm{0.95}$;$8.55\pm{6.75}$ \\
\hline
\end{tabular}
\end{center}
\end{table}
% Root-mean-square (RMS) amplitude values describe the loudness for each recording environment.
% \vspace{-1mm}
\subsection{Data Preprocessing}
The two different recording environments also resulted in different levels of clipping. As shown in Table~\ref{tab:data2}, clipping occurs much more often in the meeting room interview recordings than the phone interviews, with an average of 10.28\% of the audio signals being clipped from their maximum range. We have reduced the level of clipping by extrapolating the clipped parts of the audio using an open-source digital audio editor \textit{Audacity}\footnote{https://www.audacityteam.org/}. In order to alleviate the effect of loudness difference shown in Table~\ref{tab:data2}, we scale the audio signal for each speaker turn separately, and make sure each turn is in the range of -1 and 1. We apply a domain-adversarial neural network based voice activity detection model \cite{Lavechin2020} for intra-speaker-turn segmentation. We also exclude speaker turns that are shorter than 2 seconds to extract robust acoustic features.

\section{Feature Extraction}
\label{sec:features-ex}
Instead of using more complex features such as mel frequency cepstral coefficients (MFCCs) or high-dimensional embeddings from pretrained speech encoders, we identify a set of acoustic and (non-verbal) turn-taking behaviour related features for their interpretability.

\textbf{Prosodic features}:
During these dyadic interviews, participants often exhibited changes in prosodic variables such as the tone and intonation of their speech when they are recollecting their experience especially from the intensive week of the study.  We compute Legendre polynomial coefficients for \textit{fundamental frequency} ($\textup{F}_0$) and \textit{energy contours}. Together with \textit{segment duration} they form 13 dynamic features to capture the prosodic variations. Prosodic features have been proven effective in identifying mood states in BD \cite{low2020automated,faurholt2016voice}.
% extract 13 \textit{fundamental frequency} ($\textup{F}_0$), \textit{energy contour}\footnote{We approximate the $\textup{F}_0$ and energy contours in each segment by Legendre polynomial expansions.} and \textit{duration} based dynamic features to capture these variations. Prosodic features have been proven effective in identifying mood states in BD \cite{low2020automated,faurholt2016voice}. 

\textbf{Rhythm features}:
We also extract 7 rhythm features from each speaker-turn using algorithm proposed by Tilsen and Arvaniti \cite{tilsen2013speech} that is based on empirical mode decomposition of the vocalic energy amplitude envelope. The envelope is decomposed into two intrinsic mode functions (IMF). They show the IMF-based rhythm features can capture information about periodicities that likely correspond to different linguistic constructs, and thus are useful for examining rhythmicity in speech. Previous studies also used them for mood state detection \cite{gideon2016mood,aldeneh2019identifying}. 

\textbf{Phonation features}:
Voice quality has an important role in signalling paralinguistic information \cite{campbell2003voice}, and previous study showed significant difference in the speaker's voice quality when comparing people who suffer from psychological disorder to healthy controls \cite{scherer2013investigating}. We compute 7 phonation-based dynamic features from sustained vowels and continuous speech utterances, including \textit{jitter}, \textit{shimmer}, \textit{amplitude perturbation quotient} and \textit{pitch perturbation quotient}. 

% \textbf{Phonological features}:
% The phonological feature system allows the phonemes to be grouped into natural classes by a set of distinctive features, which gives insight into how classes of sounds pattern together when undergoing various phonological processes \cite{chomsky1968sound}. These phonological features are commonly understood by clinicians and are used to characterise pathological speech \cite{jiao2017interpretable,vasquez2019phonet}. We adopt the feature learning model proposed in \cite{vasquez2019phonet} that trains a bank of bidirectional RNNs for estimating the posterior probabilities of the occurrence of 18 phonological classes. These classes include \textit{consonantal}, \textit{continuant}, \textit{Labial}, \textit{nasal}, \textit{anterior}, \textit{voiced} etc.

\textbf{Dialogue features}:
To model high-level interactive patterns in the dialogue, we extract the set of 13 turn-taking behaviour related features following the work in \cite{aldeneh2019identifying,wang2020learning}, including \textit{relative floor control}, \textit{turn hold offset}, \textit{number of consecutive turns}, \textit{turn switch offset}, \textit{speech overlaps} and \textit{number of words per second}, per turn. 

All acoustic and dialogue features are Z-normalised either: 1) using the mean and standard deviation of each interview respectively, denoted as ``Person'', or 2) using the mean and standard deviation of all training samples, denoted as ``Global''. Previous studies \cite{gideon2016mood,cummins2011investigation} suggest normalisation by ``Person'' focuses on each individual baseline and can reduce the potential extraneous effect, for example, caused by different recording environment. 

% \vspace{-4mm}
\section{Feature Aggregation}
\label{sec:features-aggr}
The paralinguistic properties drawn from a patient's response in an interview, represented by the features described in Section~\ref{sec:features-ex}, can be indicative of the state of the patient at the time of speaking. However, the state of the patient also evolves while being affected by the highly dynamic nature of a dyadic conversation. The common approach of applying high-level statistical functions (HSFs) does not effectively capture this dynamics. In this section we describe the method of signature transform (SIG), which we use to aggregate frame-level acoustic or turn-level dialogue features and encode nonlinear time-dependent interactions in the feature set, $S(X)$.

A sequentially ordered data stream may be thought of as a discretisation of a path of finite length $X:[a,b]\to \mathbb{R}^d$, where $a\leq b$ and $d\in \mathbb{N}$. For example, a speech sequence represented by its phonation-based dynamic features can be thought of as a path of $d=7$. \textit{Signature transform}, also known as \textit{path signature}, describes a graded sequence of statistics characterising the underlying path \cite{chen1958integration,lyons2007differential}, and thus provides an effective feature set for capturing its total ordering (i.e., incremental effect) while ignoring the positional effect. More specifically, consider a $d$-dimensional path $X$ over the time interval $[a,b]$, the signature $S(X)$ of this path is the infinite collection of statistics\footnote{We refer the readers to \cite{kidger2019deep} Appendix A for a formal definition of signature transform, and \cite{chevyrev2016primer} for a primer on its use in machine learning.}:
\begin{equation*}
\begin{aligned}
    % S(X)_{a,b} = (1, S(X)_{a,b}^1, \dots, S(X)_{a,b}^d, S(X)_{a,b}^{1,1}, S(X)_{a,b}^{1,2}, \dots)
    S(X)_{a,b} = \left ( \left \{ S(X)_{a,b}^{i_1} \right \}_{1\leq i_1\leq d}, \left \{ S(X)_{a,b}^{i_1,i_2} \right \}_{1\leq i_1, i_2\leq d}, \dots \right )
\end{aligned}
\end{equation*}
where each term is a $n$-fold iterated integral of $\textup{x}$ with multi-index $i_1, \dots, i_k$:
\begin{equation*}
    S(X)_{a,b}^{i_1, \dots, i_k} = \underset{\underset{t_{1}, \dots, t_{k} \in [a, b]}{t_{1} < \dots < t_{k}}} { \int \dots \int} dX_{t_{1}}^{i_1} \otimes \dots \otimes dX_{t_{k}}^{i_k}
\end{equation*}
where $k \in \mathbb{N}$. $S(X)_{a,b}^{i_1, \dots, i_k}$ is termed as the $k$th level of the signature. In practice we truncate the signature to order $n$ to have a finite dimensional representation. Note the positional information in the data stream can be informative in many applications. For example, it's useful to know the beginning and end of each speaker turn when modelling the conversation.
% where the degree of its iterated integrals is no greater than $n$. This ensures the path signature has finite dimensional representation. Motivated by its ability to naturally capture sequential ordering, we apply signature transform (SIG) to frame-level acoustic and turn-level dialogue features ($X$), to generate interview-level fixed-length representation $S(X)$.
% Let $TS(X)_{a,b}^n$ denote the truncated signature of $X$ of order $n$, i.e.
% \begin{equation*} O9288JzNvDwa
%     TS(X)_{a,b}^n = (1, S(X)_{a,b}^1, \dots, S(X)_{a,b}^{k_{n}})
% \end{equation*}
% The dimensionality of the truncated path signature is $(d^{n+1}-d)(d-1)^{-1}$. 
The visibility transform, initially introduced in \cite{yang2017leveraging}, is able to embed the effect of the absolute positions of the input sequence into signature transform, which is otherwise not included in the signature as it is translation-invariant map\footnote{The mathematical definition of the visibility transform can be found in \cite{wu2020signature}.}.

Consider a $d$-dimensional data sequence of length $n$, i.e., $\textup{x}= \left ( x_1,\dots, x_n \right )$.  The visibility transform in prior to computing signature, adds two time steps and a binary coordinate to the input sequence $\textup{x}$ that is equal to 1 until the second-to-last time step:
% \begin{equation*}
% % \begin{aligned}
%   \phi \left ( \textup{x} \right )= \left ( \left (1, x_1 \right ),\dots, \left (1, x_{n-1} \right ), \left (1, x_{n} \right ), \left (0, x_{n} \right ), \left ( 0, 0 \right ) \right ),
% % \end{aligned}
% \end{equation*}
\begin{equation*}
% \begin{aligned}
  \phi \left ( \textup{x} \right )= \left ( \mbox{ap}_1(x_1),\dots, \mbox{ap}_1( x_{n-1}), \mbox{ap}_1( x_{n}), \mbox{ap}_0(x_{n}), \mathbf{0} \right ),
% \end{aligned}
\end{equation*}
where $\mbox{ap}_c:\mathbb{R}^d\to \mathbb{R}^{d+1}$ is an operator expanding a $d$-dimensional vector to $d+1$ dimensions by appending scalar $c$ at the end, and $\mathbf{0}\in \mathbb{R}^{d+1}$. The resulting sequence  $\phi \left ( \textup{x} \right )$ is $d+1$-dimensional and of length $n+2$.  We apply visibility transform to our frame-level acoustic features per speaker-turn before feature aggregation. This way the start and end positions of each turn can be embedded in the interview-level signature representation. We name this approach as VT-SIG.
% Considering we have several sequences of frame-level acoustic features per speaker-turn, we experiment with augmenting each with visibility transform before feature aggregation with signature, so the start and end of each turn can be embedded in the interview-level signature representation. We name this approach as VT-SIG.

% Each interview can be thought of as a series of verbal and non-verbal messages expressed in the speech signal, by each or both of the speakers. Although the paralinguistic function of these messages can be indicative of the state and traits of the speaker, and such state is also evolutionary that is affected by the context in the real dynamic interaction scenario. 

\section{Experiments and Analysis}
\label{sec:exp}
Following previous work we choose nested leave-one-subject-out as the evaluation scheme, and logistic regression with L2 regularisation for classification. For each fold, we first apply VT-SIG to each feature type, and keep only the first three levels of the path signature\footnote{We use iisignature Python library, \url{https://pypi.org/project/iisignature/}.}. We conduct feature selection on signature-transformed interview-level features through computing Pearson Correlation Coefficients (PCC) with the IPDE scores\footnote{The International Personality Disorder Examination (IPDE) \cite{loranger1997assessment} is a semi-structured clinical interview designed to assess major categories of personality disorders.} on the training data and retain the features with $p$-values less than 0.001. This results in a small number of features. The selected features are then fed to the classifier for 3 separate binary tasks: (1), BD vs. healthy controls, (2), BPD vs. healthy controls, and (3), BD vs. BPD patients. 
% We conduct three separate experiments, extracting features from the speech of each participant and interviewer respectively, as well as the whole interview (as a sequence of turns) without speaker identification (denoted as `Both')\footnote{For the `Interviewer' and `Both' experiments, we increase the p value threshold to 0.002.}. 

% \subsection{Correlation study of features}
% \label{sec:corr}
% Mixed effect modelling... not ready

% \begin{table}[htb]
% \centering
% \scalebox{0.9}{
% \begin{tabular}{ccc}
% \hline
% \multicolumn{1}{c}{} & \multicolumn{2}{c}{\textbf{BD vs. H}} \\
% \bf Feature & \bf Estimate & \bf $p$-value \\ \hline
% (DEPID, MATTR, BI) & 18.722$\pm$8.139 & * \\
% (Nonflu., Verbs) & 8.022$\pm$2.887 & ** \\
% \hline 
% \end{tabular}}
% \caption{Features with coefficients and standard errors from GLMs. The main coefficients indicate differentiating between BD patients and healthy controls. We report $p$-values obtained from likelihood ratio test against a null model with no diagnosis prediction effect. \textit{$p$-value codes: `**'$<$ 0.01, `*'$<$ 0.05}}
% \label{tab:analysis1}
% \end{table}

\subsection{Analysis of the selected features}
Five most significant and commonly selected features from each task are briefly summarised in Table~\ref{tab:feat-ranking} as examples. Each interview-level feature (a signature term) is represented as a linear combination of the original frame-level or turn-level features. We see almost all of the selected features are volume integrals, i.e., they are triple integrals of three acoustic/dialogue features. We notice the importance of the binary coordinate (i.e., \textit{c} in Table~\ref{tab:feat-ranking}) from visibility transform, to the task of \textit{H vs. BD}. For example, \textit{(c, apq, logE)} represents the nonlinear effect between \textit{apq} and \textit{logE} only at the last frame of each turn of speech\footnote{If \textit{c} is indexed first in the signature term, then this term only captures the effect from the coordinates of the following indices at the last time step. This is proved in Theorem 5 of \cite{wu2020signature}.} while its feature values before the last frame are shown to be unimportant. \textit{(IMF1\_m, c, c)} and \textit{(c, IMF1\_m)}, which represent the linear incremental effect of \textit{IMF1\_m} and the linear effect of \textit{IMF1\_m} at the last frame (of each turn) respectively, are also among the selected features. 

While most of the important features for \textit{H vs. BD} represent either purely incremental or (last-)positional effects, the selected signature terms for the other two tasks are of nonlinear mixed effects from both aspects. Overall, we notice the importance of the phonation features in two tasks involving BD and in particular, \textit{BD vs BPD}. Rhythm features, which have previously used for mood state detection, are selected in distinguishing between the BD/BPD patients and healthy controls. For each acoustic/dialogue feature, we also analyse its potential interaction effect with the recording environment, who the interviewer is and the gender of the participant, which we treat as control variables, by fitting a linear model with diagnosis as response. We find the `recording environment' variable significantly changed the effect of \textit{(apq, c, logE)} ($p<0.05$). As a result we remove it from the final feature set for classification, this way we only model effects that are not systematically variable across situational factors.
% , and the appearances of the dialogue features in the task of \textit{H vs BPD}. Rhythm features, which have previously used for mood state detection, are selected in distinguishing between the BD/BPD patients and healthy controls. 

% For each speech/dialogue feature, we analyse its potential interaction effect with the recording environment, who the interviewer is and the gender of the participant, which we treat as control variables, by fitting a linear model with diagnosis as response. If we observe a significant interaction between the feature and any of the control variables ($p<0.05$) then we remove it from the final feature set for classification. For example, we removed \textit{(apq, c, logE)} from the selection as the `interviewer' variable has shown to change the effect of this feature on the diagnosis outcome.
\vspace{-2mm}
\begin{table}[htb]
\begin{center}
\caption[dummy caption]{Top-5 most significant and commonly selected features during LOOCV. Features belong to the dialogue category are colored in {\color{red}red}; rhythm features are in {\color{green}green} and phonation features are in {\color{blue}blue}.\footnotemark[7]}
\label{tab:feat-ranking}
\vspace{0.1cm}
\scalebox{0.85}{
\begin{tabular}{c|c|c}
\bf H vs BD & \bf H vs BPD & \bf BD vs BPD \\ \hline
\color{blue}\textit{(c, apq, logE)} & \color{green}\textit{(SPBr, IMF12, IMF2\_m)} & \color{blue}\textit{(DF0, Jitter, logE)} \\
\color{blue}\textit{(c, logE, apq)} & \color{green}\textit{(CNTR, IMF12, IMF2\_m)} & \color{blue}\textit{(logE, DDF0, Jitter)} \\
\color{blue}\textit{(apq, c, logE)} & \color{red}\textit{(TL, n\_OVL, TSO)} & \color{blue}\textit{(logE, DF0, Jitter)} \\
\color{green}\textit{(c, IMF1\_m)} & \color{red}\textit{(n\_OVL, RFC\_t, SP\_m)} & \color{blue}\textit{(logE, Jitter, DDF0)} \\
\color{green}\textit{(IMF1\_m, c, c)} & \color{red}\textit{(n\_OVL, RFC\_t, n\_LP)} & \color{blue}\textit{(ppq, logE, DF0)} \\
\hline
\end{tabular}}
\end{center}
\end{table}
\footnotetext[7]{\textit{c}: binary coordinate added during visibility transform; \textit{apq}: amplitude perturbation quotient, \textit{logE}: logaritmic energy; \textit{DF0/DDF0}: first/second derivative of the fundamental Frequency; \textit{IMF1\_m}: mean within-utterance instantaneous freq. of IMF1; \textit{IMF12}: ratio between IMF2 and IMF1; \textit{SPBr}: ratio between power in envelope spectrum bands (1/3.5/10 Hz); \textit{CNTR}: envelope spectrum centroid computed over 1-10 Hz band; \textit{TL}: duration of each speaker turn; \textit{n\_OVL}: number of speech overlaps; \textit{TSO}: latency between speaker turn transitions; \textit{RFC\_t}: relative floor control (time); \textit{SP\_m}: average length of short pauses; \textit{n\_LP}: number of long pauses ($>$500 ms).}

% \subsection{Evaluation of reprocessing}

\vspace{-4mm}
\subsection{Results and Discussion}
\label{sec:res}
We first summarise the classification results obtained by using visibility transform enhanced signature (VT-SIG) in Table~\ref{tab:results}. With the (late) fusion of acoustic and dialogue features extracted from participants' speech only, we obtain a AUROC of 0.717 in H/BD, 0.841 in H/BPD and 0.716 in BD/BPD. The classification performance drop sharply when we switch to interviewers' speech instead, due to the non-clinical nature of the interviews. Modelling the interviews as a sequence of turns from both speakers (named ``Both'') also result in worse performance than learning from only the participants.
% \footnote{We also report a AUROC of 0.0 (95\% CI 0.0-0.0) in H/BD, 0.0 (95\% CI 0.0-0.0) in H/BPD and 0.0 (95\% CI 0.0-0.0) in BD/BPD, obtained from running 10-fold cross-validation for 10 times, for measuring model instability.}
% \textit{As the purpose of the interviews were merely to understand the individual's experience of taking part in the AMoSS study rather than establishing their mental state at the time of interview, it is no surprise that features extracted from the interviewers have very weak discriminative power. Modelling the interviews as a sequence of utterances also resulted in much worse performance than learning from the participants alone.}
\vspace{-2mm}
\begin{table}[htb]
\centering
\caption{Classification results for three binary tasks: H vs. BD, H vs. BPD and BD vs. BDP, using logistic regression. Results shown are macro-averaged $\mathrm{F_1}$ and AUROC scores across all interviews. $p$-value used for feature selection: `*'$<$ 0.005, or else $<$ 0.001 is used.}
\label{tab:results}
\vspace{0.2cm}
\scalebox{0.85}{
\begin{tabular}{c|cccccc}
\hline
\multicolumn{1}{c}{} & \multicolumn{2}{c}{\bf H/BD} & \multicolumn{2}{c}{\bf H/BPD} & \multicolumn{2}{c}{\bf BD/BPD} \\
\bf Subject & $\mathbf{F_1}$ & \bf AUC & $\mathbf{F_1}$ & \bf AUC & $\mathbf{F_1}$ & \bf AUC \\ \hline
Participant & \textbf{0.738} & \textbf{0.738} & \textbf{0.827} & \textbf{0.841} & \textbf{0.710} & \textbf{0.716}\\\
Interviewer & 0.581 & 0.583 & 0.102* & 0.100* & 0.552* & 0.556*\\
Both & 0.477 & 0.488 & 0.512* & 0.515* & 0.683 & 0.686\\
\hline 
\end{tabular}}
\end{table}

We also compare results obtained by using different feature normalisation and aggregation methods described in Section~\ref{sec:features-ex} and~\ref{sec:features-aggr}. First, we notice VT-SIG in general has the better performance and is more reliable across all three tasks. As for HSF though we have to increase the p-value feature selection threshold from 0.001 to 0.005 or even 0.01 to have any feature for H/BD and H/BPD, it still obtains relatively poor performance. Secondly, it is also shown normalising features per interview (``Person'') has led to increased performance for signature-based models with and without the use of visibility transform (namely, VT-SIG and SIG). Different feature normalisation methods have not made any significant impact on the overall performance of the HSF-based models.
\vspace{-2mm}
\begin{table}[htb]
\centering
\caption[dummy caption]{Performance comparison among different feature aggregation (Aggr)\footnotemark[8] and normalisation methods (Norm). Results shown are average \textbf{AUROC}s across all interviews. $p$-value codes: `**'$<$0.001; `*'$<$0.005; `$\dagger$'$<$0.01.}
\label{tab:full-results}
\vspace{0.2cm}
\scalebox{0.95}{
\begin{tabular}{c|c|ccc}
\hline
% \multicolumn{1}{c}{} & \multicolumn{1}{c}{} & \multicolumn{3}{c}{\bf AUROC} \\
\bf Aggr & \bf Norm & \bf H/BD & \bf H/BPD & \bf BD/BPD\\ \hline
HSF & None & 0.381$^{\dagger}$ & 0.544$^{\dagger}$ & 0.681**\\
HSF & Global & 0.405$^{\dagger}$ & 0.515$^{\dagger}$ & 0.686**\\
HSF & Person & 0.575* & 0.556* & 0.543**\\
SIG & None & 0.554* & 0.699** & 0.644**\\
SIG & Global & 0.435* & 0.699** & 0.662**\\
SIG & Person & 0.661* & 0.686** & 0.716**\\
VT-SIG & None & 0.608** & 0.728** & 0.710**\\
VT-SIG & Global & 0.554** & 0.811** & 0.633**\\
VT-SIG & Person & \textbf{0.738**} & \textbf{0.841**} & \textbf{0.716**}\\
\hline 
\end{tabular}}
\end{table}
\footnotetext[8]{For reasonable comparison we adopt a wide range of HSFs previously used in \cite{gideon2016mood}.}

\vspace{-2mm}
\section{Conclusions and Future Work}
% In this paper, we demonstrate the potential of using speech from non-clinical interviews to the automatic assessment of BD and BPD, which is challenging for clinicians to distinguish. 
% While most existing studies learn from clinical interviews to automatically screen mental health conditions, the detection of BD and BPD is still understudied. In this paper, we demonstrate the potential of using speech from non-clinical interviews to the automatic assessment of two conditions. 
In this paper, we demonstrate the potential of using speech from non-clinical interviews for detecting BD and BPD. Modelling short-term features and generating final representation is key for any machine learning based mental health assessment model. We propose the use of visibility-signature transform that embeds sequential ordering for feature aggregation. We show the better performance obtained by the proposed approach comparing with widely used high-level statistical functions. For future work, we plan for new data collection with more participants and multiple interviews per subject, from two different locations, which will allow for longitudinal studies and cross-site validation.

\section{Acknowledgements}
\vspace{-2mm}
This work was supported by the MRC Mental Health Data Pathfinder award to the University of Oxford [MC\_PC\_17215], by the NIHR Oxford Health Biomedical Research Centre and by the The Alan Turing Institute under the EPSRC grant EP/N510129/1. We would also like to thank John Gideon and Fasih Haider for sharing code and thoughts. 
% We would also like to thank John Gideon and Fasih Haider for sharing code and thoughts.
% The views expressed are those of the authors and not necessarily those of the NHS, NIHR or the Department of Health.

% References should be produced using the bibtex program from suitable
% BiBTeX files (here: strings, refs, manuals). The IEEEbib.bst bibliography
% style file from IEEE produces unsorted bibliography list.
% -------------------------------------------------------------------------
\bibliographystyle{IEEEbib}
\bibliography{refs}

\begin{thebibliography}{10}

\bibitem{coryell1993enduring}
W. Coryell, W. Scheftner, M. Keller, J. Endicott, J. Maser, and G.~L. Klerman,
\newblock ``The enduring psychosocial consequences of mania and depression,''
\newblock {\em The American journal of psychiatry}, 1993.

\bibitem{goldberg1995course}
J.~F. Goldberg, M. Harrow, and L.~S. Grossman,
\newblock ``Course and outcome in bipolar affective disorder: a longitudinal
  follow-up study,''
\newblock {\em The American journal of psychiatry}, 1995.

\bibitem{zimmerman2015psychosocial}
M. Zimmerman, W. Ellison, T.~A. Morgan, D. Young, I. Chelminski, and K.
  Dalrymple,
\newblock ``Psychosocial morbidity associated with bipolar disorder and
  borderline personality disorder in psychiatric out-patients: comparative
  study,''
\newblock {\em The British Journal of Psychiatry}, 2015.

\bibitem{zimmerman2013relationship}
M. Zimmerman and T.~A. Morgan,
\newblock ``The relationship between borderline personality disorder and
  bipolar disorder,''
\newblock {\em Dialogues in clinical neuroscience}, 2013.

\bibitem{zimmerman2013distinguishing}
M. Zimmerman, J.~H. Martinez, T.~A. Morgan, D. Young, I. Chelminski, and K.
  Dalrymple,
\newblock ``Distinguishing bipolar ii depression from major depressive disorder
  with comorbid borderline personality disorder: demographic, clinical, and
  family history differences,''
\newblock {\em The Journal of clinical psychiatry}, 2013.

\bibitem{low2020automated}
D.~M. Low, K.~H. Bentley, and S.~S. Ghosh,
\newblock ``Automated assessment of psychiatric disorders using speech: A
  systematic review,''
\newblock {\em Laryngoscope Investigative Otolaryngology}, vol. 5, no. 1, pp.
  96--116, 2020.

\bibitem{cummins2015review}
N. Cummins, S. Scherer, J. Krajewski, S. Schnieder, J. Epps, and T.~F.
  Quatieri,
\newblock ``A review of depression and suicide risk assessment using speech
  analysis,''
\newblock {\em Speech Communication}, vol. 71, pp. 10--49, 2015.

\bibitem{parola2020voice}
A. Parola, A. Simonsen, V. Bliksted, and R. Fusaroli,
\newblock ``Voice patterns in schizophrenia: A systematic review and bayesian
  meta-analysis,''
\newblock {\em Schizophrenia Research}, vol. 216, pp. 24--40, 2020.

\bibitem{faurholt2016voice}
M. Faurholt-Jepsen, J. Busk, M. Frost, M. Vinberg, E.~M. Christensen, O.
  Winther, J.~E. Bardram, and L.~V. Kessing,
\newblock ``Voice analysis as an objective state marker in bipolar disorder,''
\newblock {\em Translational psychiatry}, 2016.

\bibitem{gideon2016mood}
J. Gideon, E.~M. Provost, and M. McInnis,
\newblock ``Mood state prediction from speech of varying acoustic quality for
  individuals with bipolar disorder,''
\newblock in {\em ICASSP}, 2016.

\bibitem{aldeneh2019identifying}
Z. Aldeneh, M. Jaiswal, M. Picheny, M. McInnis, and E.~M. Provost,
\newblock ``Identifying mood episodes using dialogue features from clinical
  interviews,''
\newblock in {\em Interspeech}, 2019.

\bibitem{matton2019into}
K. Matton, M.~G. McInnis, and E.~M. Provost,
\newblock ``Into the wild: Transitioning from recognizing mood in clinical
  interactions to personal conversations for individuals with bipolar
  disorder.,''
\newblock in {\em Interspeech}, 2019.

\bibitem{arribas2017signature}
I.~P. Arribas, K. Saunders, G. Goodwin, and T. Lyons,
\newblock ``A signature-based machine learning model for bipolar disorder and
  borderline personality disorder,''
\newblock {\em Translational Psychiatry}, p. 274, 2018.

\bibitem{wang2019path}
B. Wang, M. Liakata, H. Ni, T. Lyons, A.~J. Nevado-Holgado, and K. Saunders,
\newblock ``A path signature approach for speech emotion recognition,''
\newblock in {\em Interspeech}, 2019.

\bibitem{kidger2019deep}
P. Kidger, P. Bonnier, I.~P. Arribas, C. Salvi, and T. Lyons,
\newblock ``Deep signature transforms,''
\newblock in {\em Advances in Neural Information Processing Systems}, 2019.

\bibitem{wang2020learning}
B. Wang, Y. Wu, N. Taylor, T. Lyons, M. Liakata, A.~J. Nevado-Holgado, and
  K.~E. Saunders,
\newblock ``Learning to detect bipolar disorder and borderline personality
  disorder with language and speech in non-clinical interviews,''
\newblock in {\em Interspeech}, 2020.

\bibitem{kim2008robust}
C. Kim and R.~M. Stern,
\newblock ``Robust signal-to-noise ratio estimation based on waveform amplitude
  distribution analysis,''
\newblock in {\em Interspeech}, 2008.

\bibitem{Lavechin2020}
M. Lavechin, M.-P. Gill, R. Bousbib, H. Bredin, and L.~P. Garcia-Perera,
\newblock ``End-to-end domain-adversarial voice activity detection,''
\newblock in {\em Interspeech}, 2020.

\bibitem{tilsen2013speech}
S. Tilsen and A. Arvaniti,
\newblock ``Speech rhythm analysis with decomposition of the amplitude
  envelope: characterizing rhythmic patterns within and across languages,''
\newblock {\em The Journal of the Acoustical Society of America}, 2013.

\bibitem{campbell2003voice}
N. Campbell and P. Mokhtari,
\newblock ``Voice quality: the 4th prosodic dimension,''
\newblock in {\em 15th ICPhS}, 2003.

\bibitem{scherer2013investigating}
S. Scherer, G. Stratou, J. Gratch, and L.-P. Morency,
\newblock ``Investigating voice quality as a speaker-independent indicator of
  depression and ptsd.,''
\newblock in {\em Interspeech}, 2013.

\bibitem{cummins2011investigation}
N. Cummins, J. Epps, M. Breakspear, and R. Goecke,
\newblock ``An investigation of depressed speech detection: Features and
  normalization,''
\newblock in {\em Interspeech}, 2011.

\bibitem{chen1958integration}
K.-T. Chen,
\newblock ``Integration of paths-a faithful representation of paths by
  non-commutative formal power series,''
\newblock {\em Transactions of the American Mathematical Society}, vol. 89, no.
  2, pp. 395--407, 1958.

\bibitem{lyons2007differential}
T.~J. Lyons, M. Caruana, and T. L{\'e}vy,
\newblock {\em Differential equations driven by rough paths},
\newblock Springer, 2007.

\bibitem{chevyrev2016primer}
I. Chevyrev and A. Kormilitzin,
\newblock ``A primer on the signature method in machine learning,''
\newblock {\em arXiv preprint arXiv:1603.03788}, 2016.

\bibitem{yang2017leveraging}
W. Yang, T. Lyons, H. Ni, C. Schmid, L. Jin, and J. Chang,
\newblock ``Leveraging the path signature for skeleton-based human action
  recognition,''
\newblock {\em arXiv preprint arXiv:1707.03993}, 2017.

\bibitem{wu2020signature}
Y. Wu, H. Ni, T.~J. Lyons, and R.~L. Hudson,
\newblock ``Signature features with the visibility transformation,''
\newblock in {\em 25th International Conference on Pattern Recognition}, 2020.

\bibitem{loranger1997assessment}
A.~W. Loranger, A. Janca, and N. Sartorius,
\newblock {\em Assessment and diagnosis of personality disorders: The ICD-10
  international personality disorder examination (IPDE)},
\newblock Cambridge University Press, 1997.

\end{thebibliography}

\end{document}